\documentclass[pdflatex,sn-mathphys-num, nonatbib]{sn-jnl}

\usepackage{graphicx}%
\usepackage{multirow}%
\usepackage{amsmath,amssymb,amsfonts}%
\usepackage{amsthm}%
\usepackage{mathrsfs}%
\usepackage[title]{appendix}%
\usepackage{xcolor}%
\usepackage{textcomp}%
\usepackage{manyfoot}%
\usepackage{booktabs}%
\usepackage{hyperref}
\usepackage{algorithm}%
\usepackage{algorithmicx}%
\usepackage{algpseudocode}%
\usepackage{listings}%
\usepackage{biblatex}
\usepackage{subfigure}

\theoremstyle{thmstyleone}%
\theoremstyle{thmstyletwo}%

\theoremstyle{thmstylethree}%

\begin{document}

\title[Article Title]{IRIS: time-structured manifold projections}


\author*[1]{\fnm{Brian} \sur{Ondov}}\email{brian.ondov@yale.edu}

\author[1]{\fnm{Chia-Hsuan} \sur{Chang}}\email{chia-hsuan.chang@yale.edu}

\author[1]{\fnm{Weipeng} \sur{Zhou}}\email{weipeng.zhou@yale.edu}

\author[2]{\fnm{Xingjian} \sur{Zhang}}\email{jimmyzxj@umich.edu}

\author[1]{\fnm{Xueqing} \sur{Peng}}\email{xueqing.peng@yale.edu}

\author[2]{\fnm{Yutong} \sur{Xie}}\email{yutxie@umich.edu}

\author[1]{\fnm{Huan} \sur{He}}\email{huan.he@yale.edu}

\author[2]{\fnm{Qiaozhu} \sur{Mei}}\email{qmei@umich.edu}

\author*[1]{\fnm{Hua} \sur{Xu}}\email{hua.xu@yale.edu}

\affil[1]{\orgdiv{Department of Biomedical Informatics and Data Science}, \orgname{Yale School of Medicine}, \orgaddress{\street{333 Cedar St}, \city{New Haven}, \postcode{06510}, \state{CT}, \country{USA}}}

\affil[2]{\orgdiv{School of Information}, \orgname{University of Michigan}, \orgaddress{\street{500 S. State St}, \city{Ann Arbor}, \postcode{48109}, \state{MI}, \country{USA}}}


\abstract{High-dimensional biomedical data, such as cell-by-gene matrices, are increasingly generated temporally. However, Manifold Learning algorithms, like t-SNE and UMAP, cannot incorporate time-ordering in their layouts, obfuscating the dynamics of cell types or other classes. As a solution, we present IRIS, a new Manifold Learning algorithm that structures layouts both chronologically and by manifold topology. IRIS can visualize a wide range of dynamic biomedical data, including scRNA-seq, comparative metagenomics, and literature.
}




\maketitle


Manifold Learning algorithms, such as t-Stochastic Neighbor Embedding (t-SNE)~\cite{van2008visualizing} and Uniform Manifold Approximation and Projection (UMAP)~\cite{mcinnes2018umap},
have become indispensable tools for exploring many types of high-dimensional biomedical data. 
Perhaps the most notable application is single-cell RNA sequencing (scRNA-seq), where learned low-dimensional layouts help to group cells according to similarity of their high-dimensional gene expression levels~\cite{becht2019dimensionality, kobak2019art}.
Further applications, however, include comparative metagenomics~\cite{armstrong2022applications}, binning for assembly~\cite{sedlar2017bioinformatics}, and Science of Science~\cite{gonzalez2024landscape, he2026medviz}.

Though Manifold Learning is typically unsupervised (i.e. does not consider labels), interpreting its layouts generally requires post-hoc coloring by known classes (such as cell types)
~\cite{kobak2019art}.
Many biomedical data types, however, are also associated with time values; for example, developmental stages in scRNA-seq, ages of subjects in comparative metagenomics, or publication dates of papers. Despite these time values being potentially crucial for understanding relationships, existing Manifold Learning algorithms cannot incorporate them in layout structure, leading to alternate solutions for conveying them.

One common workaround is to color a layout by both class and by time in side-by-side views~\cite{gonzalez2024landscape, qiu2024single}.
Though this can convey some high-level trends, more complex interactions remain obfuscated by the tedious scanning between plots that is required. For example, estimating the time range of a given class is difficult and imprecise, and identifying its earliest or latest points is nearly impossible.

Another proposed solution is Aligned UMAP, which produces multiple layouts while penalizing movement of individual items across them~\cite{dadu2023application}. However, this requires (1) a small number of discrete time slices, and (2) the same items appearing in multiple slices, which is not the case for scRNA-seq data (and much other temporal data), requiring additional workarounds to induce links. 
Further, this solution only exacerbates the problem of multiple views, requiring either animation or unwieldy 3-dimensional strand plots to show trajectories.



Here, we propose structuring a \textit{single} low-dimensional layout both chronologically and by high-dimensional relationships. To achieve this, we develop IRIS, a new Manifold Learning algorithm whose inductive bias
creates circular layouts where time radiates outward in concentric rings. This makes temporal relationships clearly visible from the layout itself, leaving the color channel available for encoding class.

\begin{figure}[htb]
  \centering
  \includegraphics[width=1\linewidth]{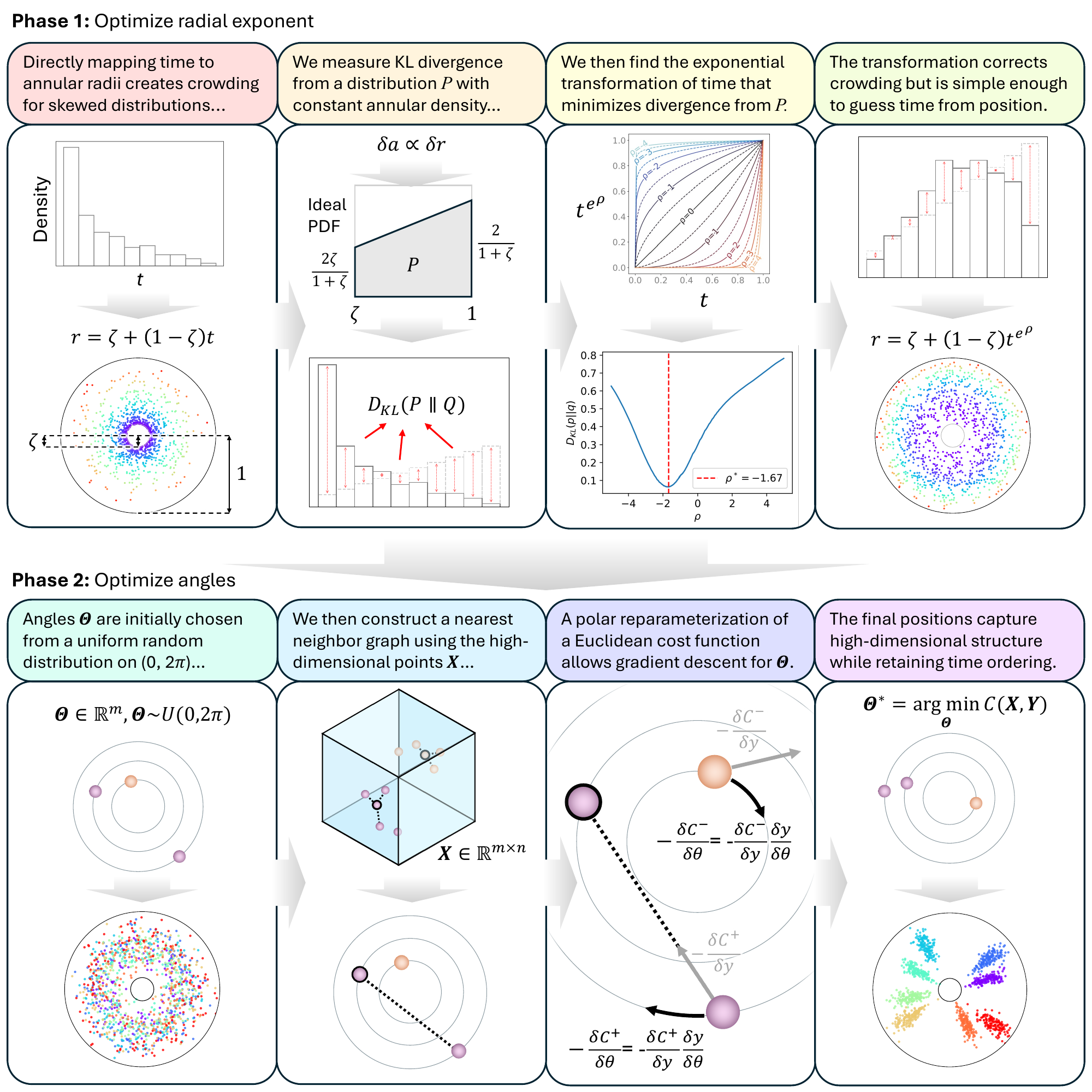}
  \caption{
  \label{fig:approach}
  Methodological overview of the IRIS algorithm. 
  }
\end{figure}

Algorithmically, IRIS operates in two main phases (Fig.~\ref{fig:approach}): (1) optimizing a global mapping of timestamp to radius, and (2) optimizing the angle of each point to minimize the divergence of distance distributions in the high- and low-dimensional spaces. The first phase accounts for temporal skewness, ensuring that density of the plot is as radially uniform as possible.
The second phase is similar to conventional Manifold Learning algorithms, but 
constrains points to radii determined in the first phase.
This is enabled by a reparameterization of a euclidean cost function from cartesian to polar coordinates.

\begin{figure}[htb!]
  \centering
  \includegraphics[width=\linewidth]{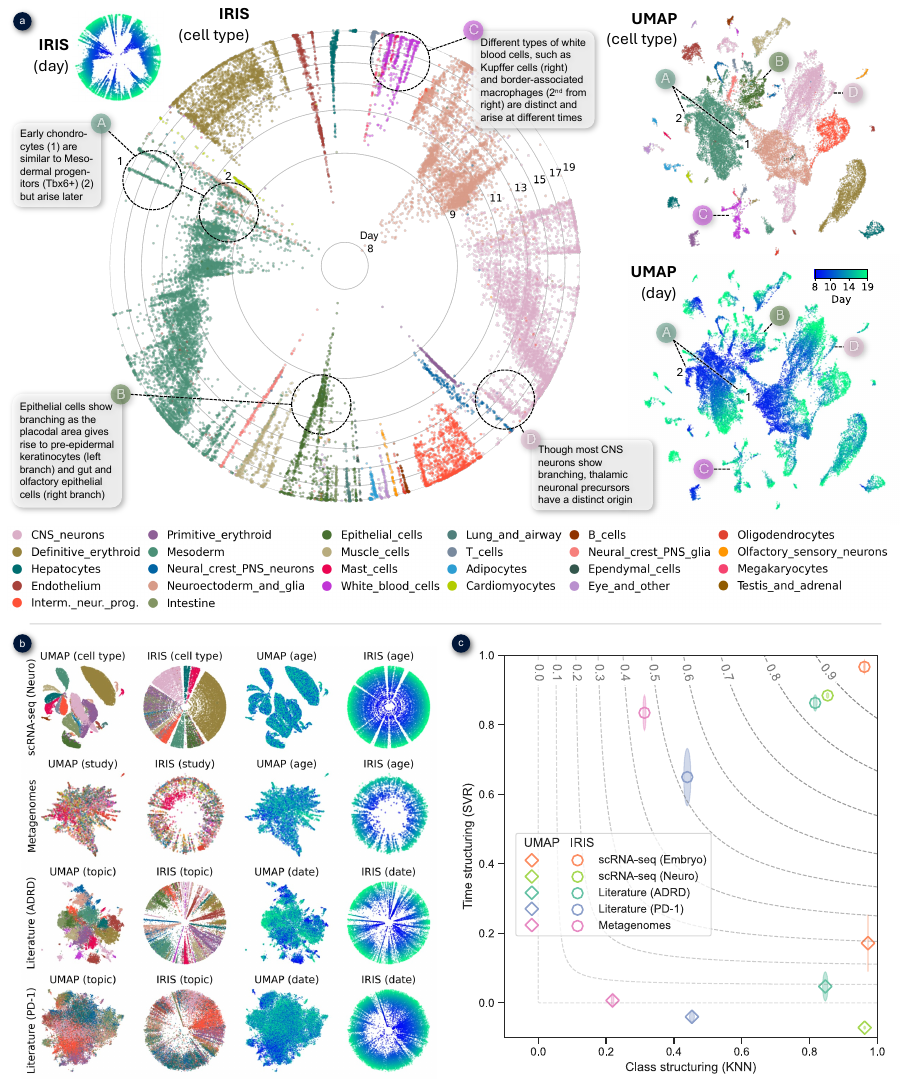}
  \caption{
  \label{fig:results}
  (a) Comparison of IRIS and UMAP  layouts for an scRNA-seq time-lapse of mouse prenatal development~\cite{qiu2024single}. 
  An interactive version of this comparison can found at \url{https://clinicalnlp.org/IRIS/iris_vs_umap.html}. (b) A similar comparison for the other four datasets.
  (c) Empirical measurement of class-structuring (via K-Nearest Neighbor classification) and time-structuring (via 2-degree polynomial kernel Support Vector Regression).
  Ellipses are dimensioned by standard deviations of each metric across five runs. Isolines are shown for the  harmonic mean of the two metrics.
  }
\end{figure}

We evaluate IRIS using five datasets spanning three diverse, high-dimensional biomedical data types: scRNA-seq, metagenomes, and transformer-based embeddings of literature (Fig.~\ref{fig:results}). Each point in these datasets is associated with both a class label (not used during learning) and an age or timestamp.
As seen in Figure~\ref{fig:results}(a), IRIS can instantly elucidate many dynamic phenomena that are not readily apparent from paired UMAP plots. Further,
quantitative metrics confirm IRIS effectively structures layouts by time (which UMAP does poorly), while retaining class structuring near that of UMAP (Fig.~\ref{fig:results}(c)).


IRIS runtime scales linearly with the number of samples (Fig.~\ref{fig:runtime}).
Future work will include improving computational efficiency, exploring joint optimization of radial and angular coordinates, and designing interactive tools specifically for the polar coordinate plots IRIS produces.
IRIS is open-source and implemented in Python and c++,\footnote{\url{https://github.com/BIDS-Xu-Lab/iris}} and we provide a NumPy-compatible Python package via PyPI.\footnote{\url{https://pypi.org/project/iris-learn/}}

\section{Methods}
\label{sec:methods}

\subsection{Algorithm and implementation}

The theoretical framework of IRIS, and the implementation reported here, are rooted in those of LargeVis~\cite{tang2016visualizing}, a forerunner of UMAP. Though their descriptions differ in mathematical framing, LargeVis and UMAP are similar algorithmically~\cite{kobak2019art}: both define graph edges based on an Approximate Nearest Neighbors search, then perform a number of iterations where sampled pairs of nodes with edges are pushed together in the layout while sampled pairs with no edges are pushed a part. We deem both the theoretical framing and the reference implementation of LargeVis to be more amenable to the time-structuring problem we study; however we note that the principles presented here could also be applied to the frameworks of UMAP, t-SNE, or other gradient-based Manifold Learning algorithms.

\subsubsection{Definitions}

Given a matrix $\mathbf{X} = \{\mathbf{x}_1,...,\mathbf{x}_m\}\in\mathbb{R}^{m\times n}$ of high-dimensional data ($m$ observations of $n$ variables) and a vector $\mathbf{t}=\{t_1,...,t_m\}\in\mathbb{R}^{m}$ of corresponding timestamps normalized to $[0, 1]$, we produce a low-dimensional layout $\mathbf{Y}=\{\mathbf{y}_1,...,\mathbf{y}_m\}\in\mathbb{R}^{m\times d}$. For our purposes, we assume $d$ is 2, though without loss of generality to 3-dimensional layouts.

\subsubsection{Desiderata}

To motivate our methods, we first define desiderata for layouts that are structured by  timestamps of data points. We note that these may not be universal and could depend on downstream uses of the layouts. However, we deem these to be reasonably broad enough to be axiomatic for defining a specific method.

\begin{enumerate}
    \item \textit{Neighbors in the layout should be similar in the high-dimensional space}. The typical core objective of nonlinear manifold learning for dimension reduction is to keep points that are close in the high-dimensional space close in the low-dimensional layout. However, as high-dimensional near neighbors may correctly move apart in a time-structured layout, we modify this desideratum to focus on a point's low-dimensional neighbors being similar (but not necessarily the most similar) in the high-dimensional space. 
    \item \textit{The timestamp of each point should be estimable.} We would like for an observer of the layout to be able to estimate the timestamp of any point given only its position in the chart. 
    \item \textit{Density should be consistent.} 
    When viewing a layout at a given scale, empty space in some areas necessarily means higher density in others, making smaller-scale relationships less clear (i.e. the `crowding problem' that t-SNE was introduced to address~\cite{van2008visualizing}). We thus desire evenness of density throughout the layout.
\end{enumerate}

\subsubsection{Choice of time mapping function}

To fulfill desiteratum (2), it is not enough to encode time into position in the plot by any arbitrarily complex function. To be truly interpretable, we will need a simple parametric function. Though this of course involves some subjectivity, we deem mapping timestamps to distance from the origin via a globally parameterized power function to be simple enough. As log transformations commonly appear in data visualizations~\cite{tufte1974data}, this should allow viewers to estimate timestamps despite the density of time changing. This mapping essentially creates a polar coordinate system where radius is a (monotonic) function of time and angle is free to be optimized. A polar formulation also comes with several benefits:

\begin{enumerate}
    \item \textit{Control of density.} In service of desideratum (3), we will need to warp time to balance heavily left- or right- skewed time distributions in our data. A power mapping of time to radius gives us broad control of the density of the entire plot, while keeping the layout space interpretable. Though this would be possible in cartesian coordinates, it may result in ``bottlenecks'' that make use of space less efficient.
    \item \textit{Continuity.} Encoding time as radius makes the optimized variable (angle) cyclic. This means any points in the layout can have near neighbors in either angular direction, potentially letting the low-dimensional layout stay more faithful to high-dimensional relationships.
    \item \textit{Conceptual distinction.} As opposed to arbitrarily choosing the $x$ or $y$ axis to represent time, giving different meanings to radius and angle may be more intuitive.
\end{enumerate}

We thus will optimize two components: (1) a global exponent of time to encode radius (rather than optimizing individual radii for each datum), and (2) the angle of each datum.
Each point $\mathbf{y}_i$ is accordingly parameterized as:

\begin{equation}
y^{(d)}_i=
\begin{cases}
r_i\cos \theta_i, & d=1 \\
r_i\sin \theta_i, & d=2
\end{cases}
\end{equation}

\begin{equation}
\label{eq:radius}
    r_i=\zeta+(1-\zeta)t_i^{e^\rho},
\end{equation}

where $\mathbf{\Theta}$ and $\rho$ are optimized and $\zeta$ is a hyperparameter (0-1) controlling the ratio of inner to outer radius (creating an annulus).

\subsubsection{Temporal resampling}
\label{sec:resampling}

In some cases, the number of distinct time labels may be small relative to the number of samples ($|\{t_i\}|<<m$) due to coarseness of estimation (for example only knowing age in years). As such sparseness can create undesirable banding artifacts when structuring a low-dimensional layout by time, IRIS offers optional probabilistic temporal resampling in service of desideratum (3). If resampling is appropriate, we propose two strategies:

\begin{itemize}
    \item \textbf{Homogenous resampling}: If time values have a known, consistent period (e.g. age or date is reported in integer years only), they can be resampled over that period ($p$):
\begin{equation}
\label{eq:sample}
    t'_i \sim U(t_i, t_i+p)
\end{equation}

    \item \textbf{Heterogeneous resampling}: If time values are on inconsistent intervals, we can assume a uniform distribution between each discrete value present in the dataset:
\begin{equation}
\label{eq:sample}
    t'_i \sim U(t_i, \min_{j;\,t_j>t_i}t_j)
\end{equation}
\end{itemize}

Resampling will, of course, trade some precision of the represented time values in order to create a more interpretable layout. We thus allow users to decide when resampling is appropriate.

\subsubsection{Optimizing the radial function}

Since time will be mapped to radius via the globally parameterized relationship in Eq.~\ref{eq:radius}, we will optimize the single parameter ($\rho$) of this relationship in an initial phase, before optimizing individual $\theta$ values for each datum with gradient descent.
To do this, we will first define an ideal distribution $P(X)$ that we would like radii of our embedded points to exhibit. Pursuant to desideratum (2), we would like to make density approximately equal for any given time value. For an increase in radius $\delta r$, the area $a$ of a circle increases proportionally ($\delta a\propto \delta r$), so $P(X)$ must increase linearly from the inner radius of $\zeta$ to the outer radius of 1. The probability density function to evenly distribute a random variable throughout a unit annulus of fractional inner radius $\zeta$ is thus:

\begin{equation}
\label{eq:pdf}
    p(x)=\begin{cases}
        0, & x<0\\
        \frac{2\zeta}{1+\zeta}+\frac{2-2\zeta}{1+\zeta}x, & 0 \leq x \leq 1 \\
        0, & x>1
    \end{cases}
\end{equation}

We wish to find the value $\rho^*$ that makes the distribution of radii $Q$ most similar to this ideal using Kullback-Leibler divergence:

\begin{equation}
\label{eq:opt-rho}
    \mathbf{\rho}^* = \arg\min_\rho D_{KL}(P \parallel Q)
\end{equation}

As a closed form of $Q$ is not available to us, we compute a probability mass function over $b$ equally sized bins for a given value of $\rho$. To avoid 0-frequency bins (which would make the KL divergence undefined), we apply Laplace smoothing with $\alpha=1$. We then accordingly discretize Eq.~\ref{eq:pdf} by integrating over each bin to obtain $P$. We can then use the discrete form of the KL divergence:

\begin{equation}
    D_{KL}(P \parallel Q) = \sum_{x \in X} P(x) \log\left(\frac{P(x)}{Q(x)}\right)
\end{equation}

Though the objective in Eq.~\ref{eq:opt-rho} is non-convex, in practice it can be easily optimized via grid search. This is due to the facts that (1) it is tractable to compute, (2) we optimize $\rho$ independently of other parameters, and (3) the composited power function in Eq.~\ref{eq:radius} ensures that a linear search of a small range of potential $\rho$ values, such as $[-5,5]$ in intervals of $0.01$, will cover both extremes while providing adequate granularity for more central values (Fig.~\ref{fig:approach}).



\subsubsection{Optimizing angles}



Given a global radial mapping, we must next optimize the angle of each datum:

\[
\mathbf{\Theta}^* = \arg\min_{\mathbf{\Theta} \in \mathbb{R}^{m}} C(\mathbf{X},\mathbf{Y}).
\]

We will base our cost function $C$ on the probabilistic framework of LargeVis~\cite{tang2016visualizing}, which defines the probability of an edge existing between any given points. This probability is maximized for $K$ near neighbors, and its inverse is maximized for sampled negative edges. To balance the desiderata, we will define both cartesian and polar components of this objective. Since we will optimize with stochastic gradient descent for pairs, we can define the positive objective $C^+$ and the negative objective $C^-$ for pairs of embeddings $(\mathbf{y_i},\mathbf{y_j})$ (positive pairs) and $(\mathbf{y_i},\mathbf{y_k})$ (negative pairs):

\begin{equation}
    C_{i,j}=C^+_{i,j}+\gamma\sum_{k=1}^KC^-_{i,k\sim P^-(\mathbf{y_k})}
\end{equation}
\begin{equation}
    C^+_{i,j}=(1-\beta)\log\frac{1}{1+(||\mathbf{y}_i-\mathbf{y}_j||^2)}+\beta\log\frac{1}{1+(\theta_i-\theta_j)^2}
\end{equation}
\begin{align}
    C^-_{i,k}=&(1-\beta)\log\left[1-\frac{1}{1+(||\mathbf{y}_i-\mathbf{y}_k||^2)}\right]+\\&\beta\log\left[1-\frac{1}{1+(\theta_i-\theta_k)^2}\right]
\end{align}

The hyperparameter $\beta$ balances euclidean and angular distances, and affects the layouts' use of angular space. We explore this further in Figures~\ref{fig:beta-results} and \ref{fig:beta-layouts}.
 To derive the gradient of these costs with respect to $\theta$, 
we first define the gradient of the cost $C$ for a pair of embeddings $(\mathbf{y}_i,\mathbf{y}_j)$ with respect to a single embeddding $\mathbf{y}_i$:

\begin{equation}
\nabla_{\mathbf{y}}C^+_{i,j} = (1-\beta)\frac{-2(\mathbf{y}_i-\mathbf{y}_j)}{1+||\mathbf{y}_i-\mathbf{y}_j||^2}
\end{equation}
\begin{equation}
\nabla_{\mathbf{y}}C^-_{i,k}=(1-\beta)\frac{2(\mathbf{y}_i-\mathbf{y}_k)}{||\mathbf{y}_i-\mathbf{y}_k||^2(1+||\mathbf{y}_i-\mathbf{y}_k||^2)}
\end{equation}

The partial derivatives of $C^+_{i,j}$ and $C^-_{i,k}$ with respect to $\theta_i$ (the angle used to define $\mathbf{y}_i$) are then:

\begin{equation}
\frac{\partial C^+_{i,j}}{\partial \theta_i} =r_i(1-\beta)\left(\frac{\partial C^+_{i,j}}{\partial y^{(2)}_i}\cos \theta_i-\frac{\partial C^+_{i,j}}{\partial y^{(1)}_i}\sin \theta_i\right)+\frac{-2\beta(\theta_i-\theta_j)}{1+(\theta_i-\theta_j)^2}
\end{equation}
and
\begin{align}
\frac{\partial C^-_{i,k}}{\partial \theta_i} =&r_i(1-\beta)\left(\frac{\partial C^-_{i,k}}{\partial y^{(2)}_i}\cos \theta_i-\frac{\partial C^-_{i,k}}{\partial y^{(1)}_i}\sin \theta_i\right)+\\ &\frac{2\beta(\theta_i-\theta_k)}{(\theta_i-\theta_k)^2(1+(\theta_i-\theta_k)^2)}.
\end{align}



Due to symmetry, $\frac{\partial C^+_{i,j}}{\partial \theta_j}=-\frac{\partial C^+_{i,j}}{\partial \theta_i}$ and $\frac{\partial C^-_{i,k}}{\partial \theta_k}=-\frac{\partial C^-_{i,k}}{\partial \theta_i}$. For each sampled pair $(\mathbf{y}_i,\mathbf{y}_j)$ and $(\mathbf{y}_i,\mathbf{y}_k)$, we can thus gain efficiency by simultaneously performing updates to $\mathbf{y}_i$, $\mathbf{y}_j$ and $\mathbf{y}_k$. 

\subsubsection{Implementation}

The first phase of IRIS is implemented purely in Python, using SciPy to compute Kullback-Leibler divergence. The second phase is based on the c++ implementation of LargeVis~\cite{tang2016visualizing}, which we refactor to parameterize optimization with angles (rather than Cartesian coordinates) and handle gradient descent updates with both polar and Cartesian components. 

\subsection{Experiments}

\subsubsection{Data}

For experiments, we choose datasets representing 3 frequently occurring types of high-dimensional biomedical data:

\begin{itemize}
    \item \textbf{scRNA-seq}: Each sample ($i$) is a cell, and each dimension ($j$) is a gene, with each value $X_{i,j}$ representing a normalized expression level. As these data tend to be highly sparse (having mostly zero values) we perform Principal Component Analysis (PCA) and use the 300 components of highest variance as input to Manifold Learning algorithms.
    \item \textbf{Metagenomes}: Each sample ($i$) is a metagenome, and each dimension ($j$) is a taxon, with each value $X_{i,j}$ representing a normalized taxonomic abundance. As abundance values are sparse, the 300 PCA components of highest variance are used as input to Manifold Learning, as for scRNA-seq data.
    \item \textbf{Literature}: Each sample ($i$) is a publication, and each $n$-dimensional representation $X_{i,1:n}$ is computed using a transformer-based embedding model~\cite{zhang2023retrieve} to encode concatenations of title and abstract, following prior work~\cite{gonzalez2024landscape, he2026medviz}. As embedding vectors are already dense, PCA is not required. Time is the date of publication, which we convert from year-month-day to a fractional year representing midnight on that date. As the day-specific distinct time values are relatively dense, we do not perform temporal resampling for literature data.
\end{itemize}

\noindent We examine five datasets across these types:

\begin{itemize}
    \item \textbf{scRNA-seq (Embryo)}: A single-cell RNA-seq dataset capturing mouse embryonic development from gastrula to birth~\cite{qiu2024single}. We use ``normalized subset 1'' from Cell X Gene,\footnote{\url{https://cellxgene.cziscience.com/e/a91d5064-aba3-4688-9fcd-6187b2f435a4.cxg/}} comprising 2,860,465 cells, and further sample \%1 of these, resulting in 28,604 cells. Samples are timestamped, for the most part, in quarter-day intervals during embryonic development, resulting in 42 discrete day values in $[8.00, 18.75]$, and at birth, which we assign a day value of 19.00. Due to the small number of discrete values relative to sample count, and the heterogeneous intervals, we apply heterogeneous temporal resampling (see \S\ref{sec:resampling}).
    The first 300 PCA components, in sum, explain 30.00\% of the variance of the original 45,525 dimensions. We use the provided \texttt{author\_major\_cell\_cluster} values as class labels for post-hoc analysis, and use \texttt{author\_cell\_type} values for further exploration of cell type arrangement within clusters.
    \item \textbf{scRNA-seq (Neuro)}: A single-cell transcriptomic atlas of neurodegenerative diseases~\cite{lee2025single}. We use the Mount Sinai NIH Neurobiobank cohort, comprising 4,140,543 samples from 1,042 individuals aged 19--89. We filter out samples with developmental stage listed as ``80 year-old and over stage'' (which is not specific enough for our purposes). As time values for IRIS, we apply homogeneous resampling to the remaining year-specific ages (see \S\ref{sec:resampling}). To make the dataset tractable for repeated Manifold Learning experiments for experimental metrics, we further sample \%1 of the remaining data, resulting in 30,556 samples. PCA reduction of 34,176 sparse gene dimensions to 300 dense components retains 96.4\% of the variance.
    Class labels are cell types as reported in Cell X Gene,\footnote{\url{https://cellxgene.cziscience.com/e/37a17b78-4864-4a42-b67b-31c00962795a.cxg/}} from which the data were downloaded.
    \item \textbf{Metagenomes (Gut)}: A collection of 4,479 gut metagenomes across 29 studies~\cite{jiang2025gutmetanet}. Time values are integer ages of subjects, resampled using the homogenous method with a period of 1 year (see \S\ref{sec:resampling}). Class labels are study of origin. PCA reduction of 999 sparse abundance columns to 300 dense components retains 99.97\% of the variance.
    \item \textbf{Literature (ADRD)}: A focused corpus of 17,518 articles related to neuroimaging assessment in Alzheimer's Disease and Related Dementias. We first search PubMed~\cite{white2020pubmed} via keyword, retrieving 230,043 titles and abstracts. We filter out pre-1975 articles (which typically lack abstracts), then perform hierarchical topic modeling with TopicForest~\cite{chang2025topicforest}, using four levels. We extract embeddings from a first-level topic labeled ``Alzheimer's Neuroimaging Assessment,'' and use TopicForest's assignment to one of 15 third-level subtopics as class labels.
    \item \textbf{Literature (PD-1)}: A focused corpus of 27,886 articles related to the immune checkpoint pathway proteins PD-1 and PD-L1. Articles are retrieved via keyword search of PubMed~\cite{white2020pubmed} using synonyms of PD-1 and PD-L1. Class labels are topic assignments from performing Latent Dirichlet Allocation~\cite{blei2003latent} with ten topics.

\end{itemize}

\subsubsection{Metrics}

To empirically assess the quality of IRIS layouts against a baseline, we define metrics for class-structure, time-structuring, and overall quality.

\begin{itemize}

\item \textbf{Class-structuring (\texttt{KNN})}: Preservation of latent high-dimensional structure (the main goal of Manifold Learning) is often measured by computing overlap of $k$ nearest neighbors in the high- and low-dimensional spaces~\cite{kobak2019art}. However, the additional constraint of time structuring makes this approach inappropriate, as high-dimensional near-neighbors may correctly be placed further away in a low-dimensional layout if their timestamps are distant. We thus instead leverage class information to measure high-dimensional structure preservation. If a manifold has been learned well, we would expect nearby points in the low-dimensional layout to be of the same class. Accordingly, we define \texttt{KNN} as the mean accuracy of a K-Nearest Neighbors (KNN) classifier predicting class from layout coordinates, using five-fold cross-validation.

\item \textbf{Time-structuring (\texttt{SVR})}: In consideration of Desideratum (2), we measure time-structuring as the ability to estimate time from layout coordinates. We thus define \texttt{SVR} as $R^2$ goodness of fit of a Support Vector Regression (SVR) with a two-degree polynomial kernel. This choice of kernel accounts for nonlinear time-structuring (such as IRIS's radial encoding), but requires the structuring to be global and relatively simple, which supports estimation by a human observer. As with \texttt{KNN}, we take the mean value from five-fold cross-validation.

\item \textbf{Time-structured Manifold Projection Score (\texttt{TMPS})}: To compute a single score capturing class- and time-structuring, we define the Time-structured Manifold Projection Score (\texttt{TMPS}) as the harmonic mean of \texttt{KNN} and \texttt{SVR}. Similarly to F1 scores, this provides a balanced composite that cannot be saturated by performance in a single dimension.

\end{itemize}

Learning for \texttt{SVR} and \texttt{KNN} metrics is performed with Scikit-learn~\cite{scikit-learn}. For experiments, we perform five replicates for each combination of dataset and algorithm. Each replicate uses different random initializations for layouts and cross-validation splits. Means and standard deviations of metrics are computed across replicates, using means from cross-validation as observations.

\subsubsection{Experimental settings}

For IRIS, we set $\zeta$ to 0.1, making the inner radius 10\% of the outer radius, and set $\beta$ to 0.95 balance toward class-structuring (see Figure~\ref{fig:beta-results}). For additional hyperparameters shared by the LargeVis framework, we set $\gamma=128$, the number of neighbors to 32, the number of steps (neighbor pairs sampled) to $m*10^3$ for each dataset size $m$, and the number of negatives samples per neighbor pair to 5. For detailed explanations of these hyperparameters and their significance, we refer readers to Tang et al.~\cite{tang2016visualizing}.
As a baseline, we compare IRIS to UMAP, which we deem to be representative of Manifold Learning algorithms in the domain of biomedicine due to its widespread adoption.
We use the originally reported UMAP implementation,\footnote{\url{https://github.com/lmcinnes/umap}} using default parameters except for \texttt{min\_dist}, which we set to 0.5, as this aids in time-structuring. An empirical comparison with the default \texttt{min\_dist} of 0.1 can be found in Figure~\ref{fig:min-dist}. All experiments were run with 10 threads using an Apple M3 Pro processor.





\bmhead{Acknowledgements}

This work was supported by the National Institute on Aging under grant number R01AG078154.





\printbibliography{}

\begin{appendices}

\section{Extended Data}\label{secA1}

\begin{figure}[htb]
  \centering
  \includegraphics[width=1\linewidth]{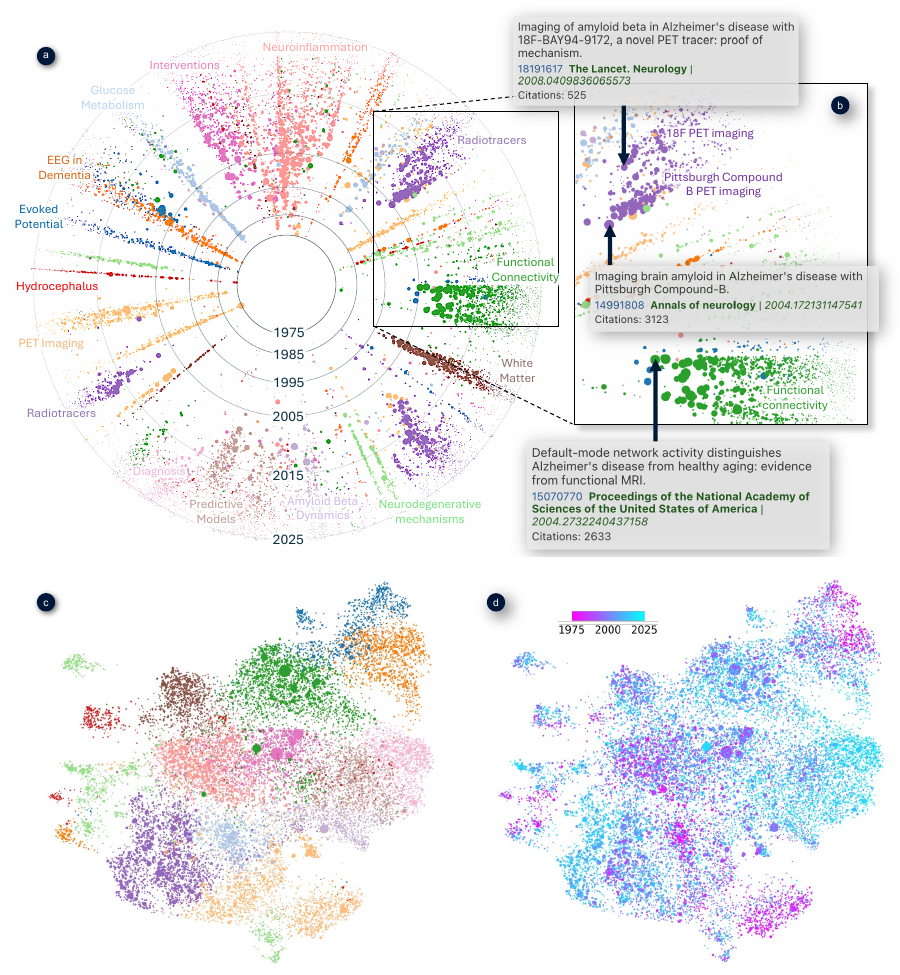}
  \caption{
  \label{fig:adrd}
Literature case study. To demonstrate the utility of IRIS for high-dimensional literature embeddings, we further explore IRIS and UMAP layouts for the Alzheimer's Disease-Related Dementias (ADRD) literature collection. Point size encodes number of citations.
(a) In addition to clearly separating independently labeled topics (as UMAP does), IRIS naturally shows the progression of each topic across the collection period (1975--2025), elucidating which topics have recently grown (e.g. radiotracers), waned (e.g. SPECT imaging) or remained consistently relevant (e.g. hydropcephalus). It further becomes clear that some topics remain isolated (e.g. evoked potential), while others split into narrower subtopics (e.g. neuroinflammation) or become admixed (e.g. diagnosis and predictive models). (b) At a more detailed level, IRIS reveals influential publications at the inceptions of subtopics. Notably, none of these phenomena are readily apparent from the UMAP layout, which requires two plots to encode (c) class and (d) time. An interactive version of the IRIS layout can be found at \url{https://medviz.org/adrd-iris}.
}
\end{figure}

\begin{figure}[htb]
  \centering
  \includegraphics[width=0.75\linewidth]{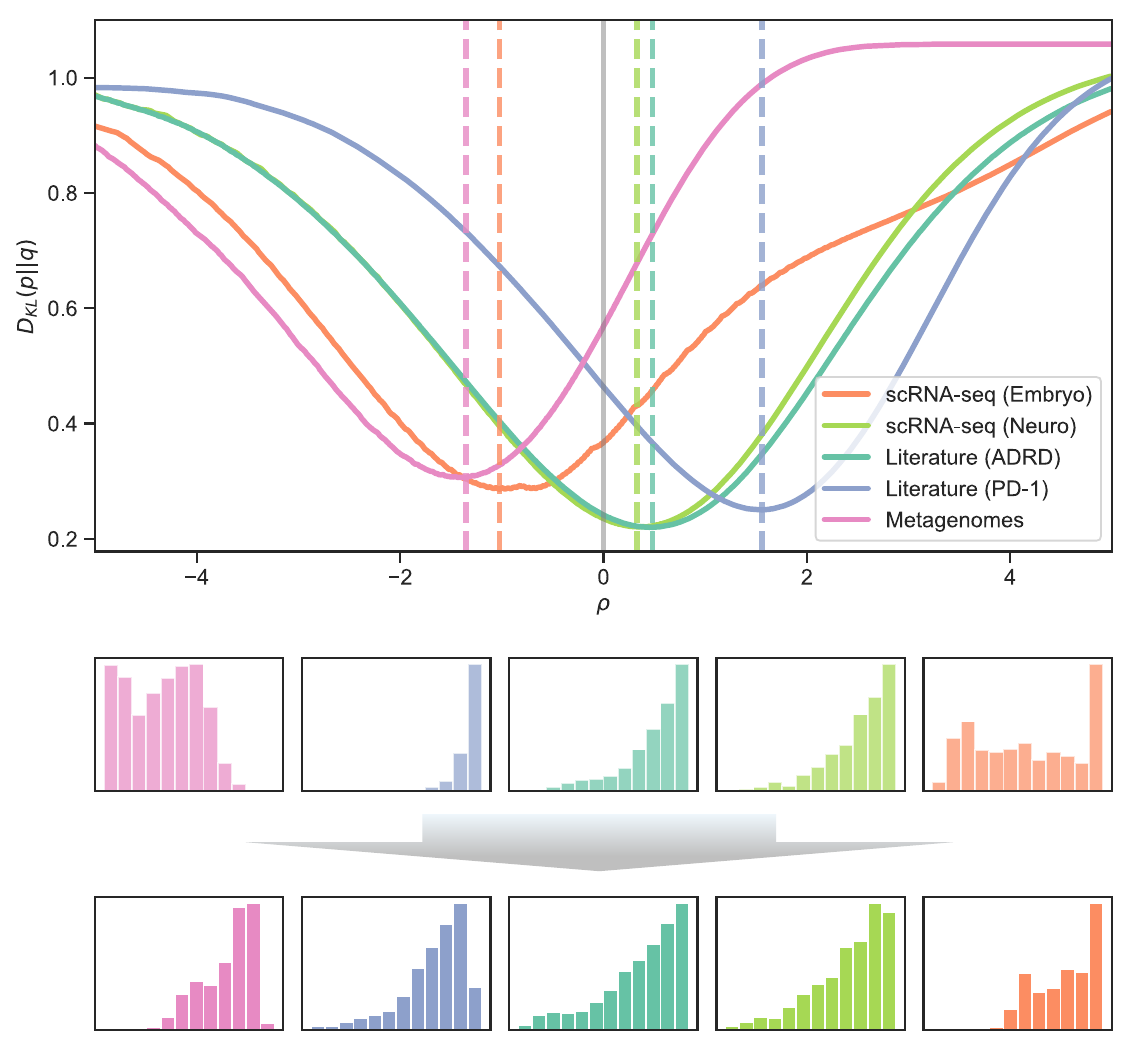}
  \caption{
  \label{fig:rho-opt}
Results of the first phases of IRIS. Top, cost curves for the five datasets within the search space of $\rho$ for the first phase of IRIS. Bottom, temporal histograms before and after optimization.
  }
\end{figure}

\begin{figure}[htb]
  \centering
  \includegraphics[width=0.75\linewidth]{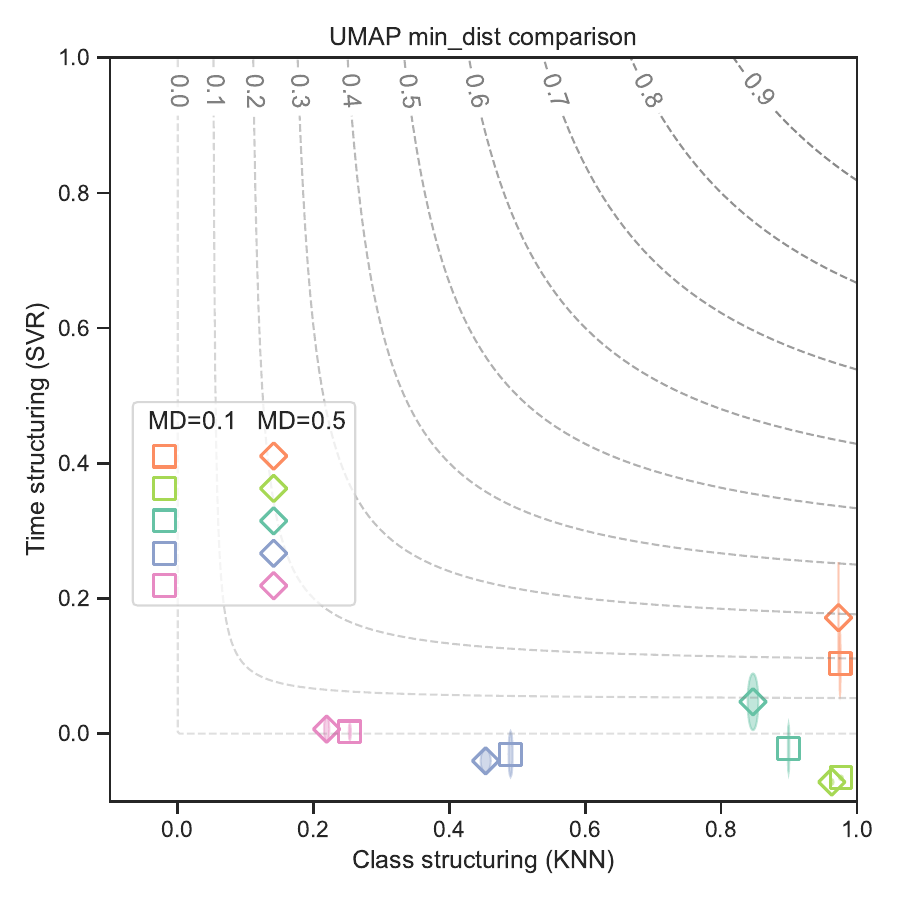}
  \caption{
  \label{fig:min-dist}
Comparison of UMAP with \texttt{min\_dist}=0.5 and its default of \texttt{min\_dist}=0.1. This hyperparameter sets the minimum distance between points in the layout, effectively controlling density. Overall, \texttt{min\_dist}=0.5 improves \texttt{TMPS} of UMAP by significantly improving \texttt{SVR} (time-structuring) for some datasets, at minor expense of \texttt{KNN}.
  }
\end{figure}

\begin{figure}[htb]
  \centering
  \includegraphics[width=.8\linewidth]{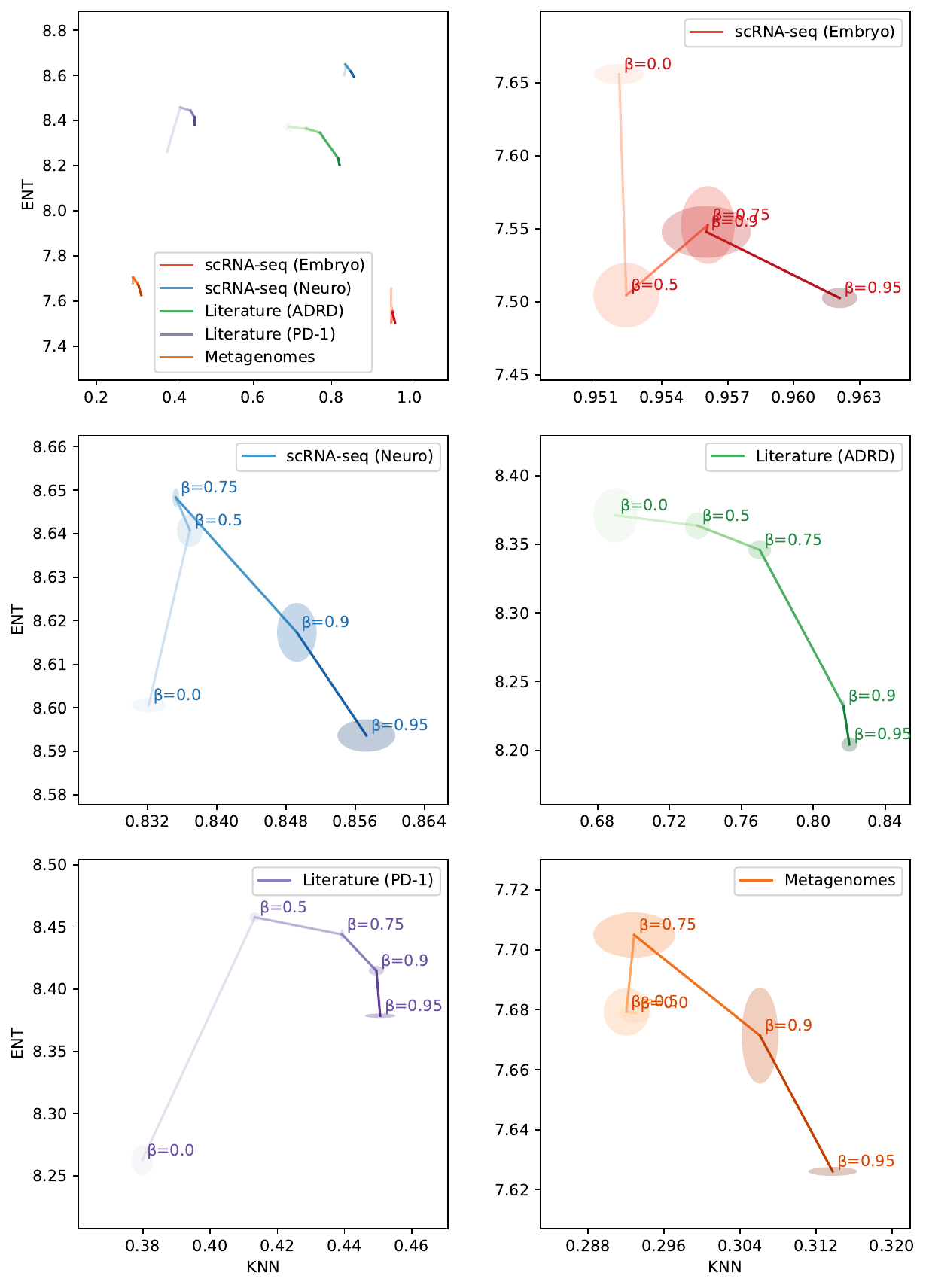}
  \caption{
  \label{fig:beta-results}
Exploration of the hyperparameter $\beta$, which determines the influences of euclidean and angular distances on the cost function. Practically, lower values of $\beta$ will allow data of different class to occupy similar angles if they were observed in different time ranges. Thus, we would expect lower $\beta$ values to improve consistency of density (desideratum (3)), especially  for datasets where classes do not stay consistently prominent over the full time range. To explore this phenomenon, we measure consistency of density by computing Shannon entropy of a two-dimensional histogram of the low-dimensional layout using 100 bins in each dimension (10,000 total). We refer to this metric as \texttt{ENT}, and higher values are better.
For each of the five datasets, we perform 3 replicates for $\beta=\{0.0, 0.5, 0.75, 0.9, 0.95\}$,
and plot \texttt{KNN} versus \texttt{ENT}. Lines in the plots get darker for increasing $\beta$ values. Vertices represent means of each metric, while widths and heights of ovals represent standard deviations for each metric.
As expected, lowering $\beta$ values generally improves class structuring but decreasing consistency of density (in other words, increasing crowding), especially for datasets whose classes are highly time-correlated, such as \textbf{scRNA-seq (Embryo)}. However, for datasets without this property, $\beta$ values that are too low become harmful (see Figure~\ref{fig:beta-layouts} for example layouts).
  }
\end{figure}

\begin{figure}[htb]
  \centering
  \includegraphics[width=\linewidth]{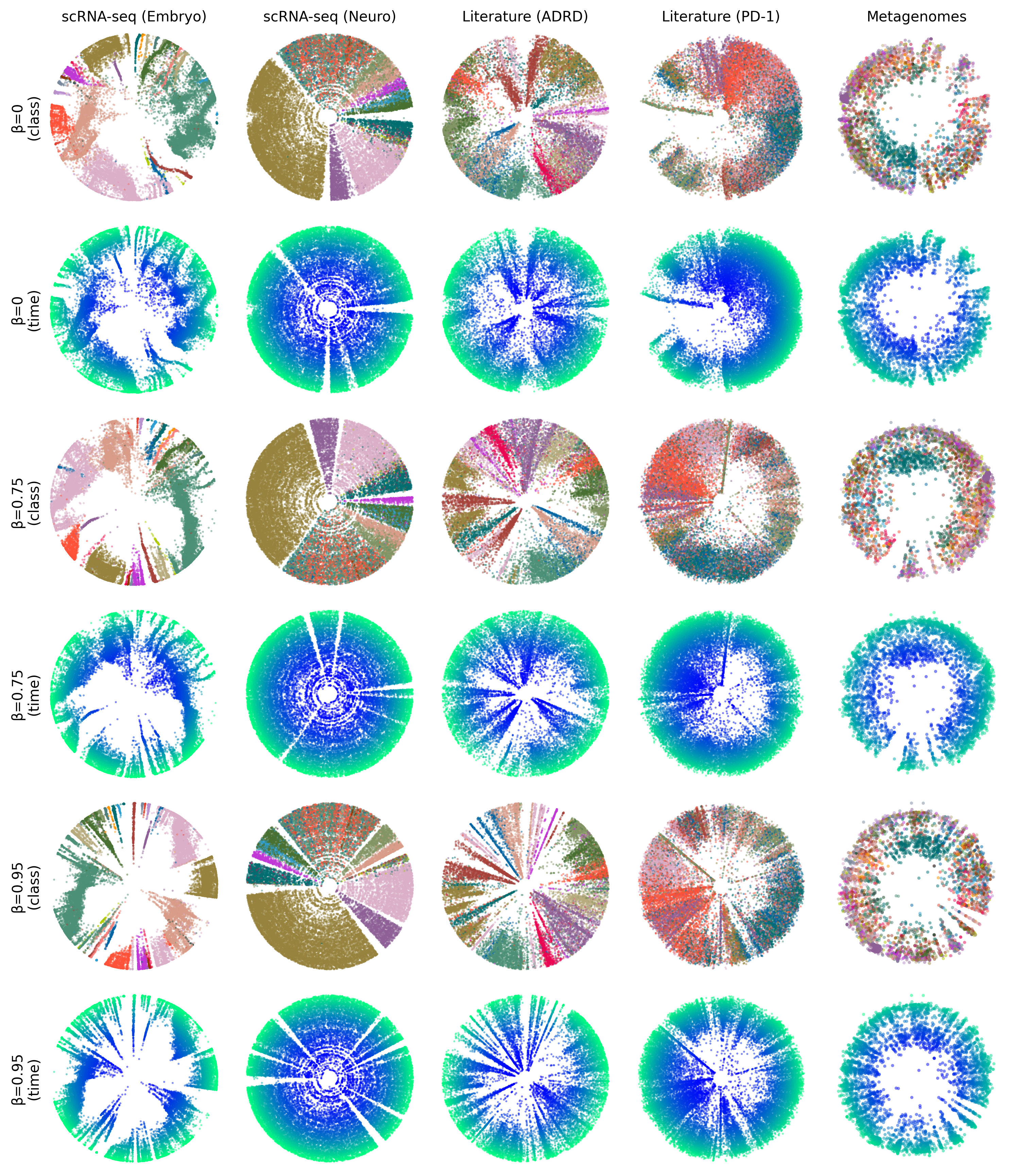}
  \caption{
  \label{fig:beta-layouts}
  Example layouts for various $\beta$ values. The influence of $\beta$ can be seen especially for the \textbf{scRNA-seq (Embryo)} dataset (first column), as classes are highly time-correlated in this set.
  }
\end{figure}

\begin{figure}[htb]
  \centering
  \includegraphics[width=0.75\linewidth]{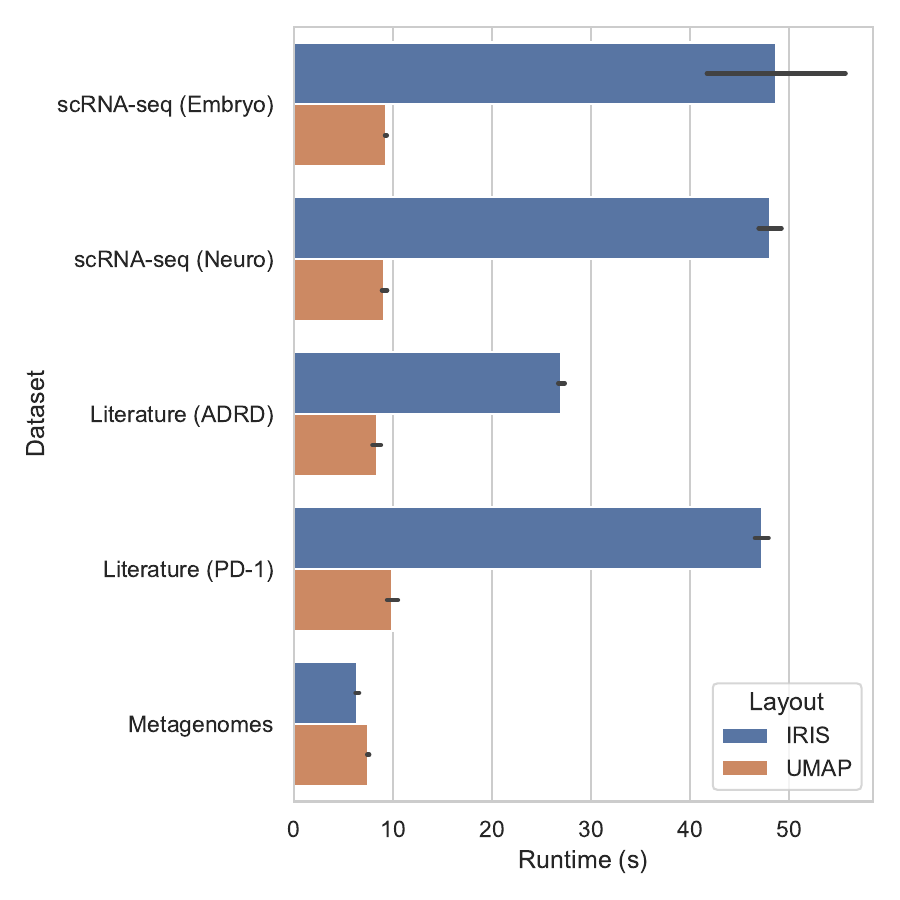}
  \caption{
  \label{fig:runtime}
  Runtimes for IRIS and UMAP across datasets. Error bars depict standard deviations for five random initializations.
  }
\end{figure}





\end{appendices}


\end{document}